\documentclass[10pt,twocolumn,letterpaper]{article}
\usepackage{iccv}

\definecolor{iccvblue}{rgb}{0.21,0.49,0.74}
\usepackage[pagebackref,breaklinks,colorlinks,allcolors=iccvblue]{hyperref}

\def\paperID{13956} 
\def\confName{ICCV}
\def\confYear{2025}

\usepackage{graphicx}
\usepackage{booktabs}

\usepackage{array}
\newcolumntype{C}{>{\centering\arraybackslash}m{0.1\linewidth}}

\usepackage[accsupp]{axessibility}  

\usepackage{orcidlink}

\usepackage{graphicx}
\usepackage{amsmath}
\usepackage{amssymb}
\usepackage{booktabs}
\usepackage{algorithm}
\usepackage{algorithmic}
\usepackage{subcaption}

\usepackage{tikz}
\usepackage[dvipsnames]{xcolor}

\newcommand{\mps}{MPS\xspace}
\newcommand{\mpss}{MPSs\xspace}

\usepackage[capitalize]{cleveref}
\crefname{section}{Sec.}{Secs.}
\Crefname{section}{Section}{Sections}
\Crefname{table}{Table}{Tables}
\crefname{table}{Tab.}{Tabs.}

\usepackage[capitalize]{cleveref}
\crefname{section}{Sec.}{Secs.}
\Crefname{section}{Section}{Sections}
\Crefname{table}{Table}{Tables}
\crefname{table}{Tab.}{Tabs.}

\newcommand{\lime}{{\sc lime}\xspace}
\newcommand{\shap}{{\sc shap}\xspace}
\newcommand{\rise}{{\sc rise}\xspace}
\newcommand{\sag}{\textsc{SAG}\xspace}

\newcommand{\gradcam}{{\sc g}rad{\sc cam}\xspace}
\newcommand{\rex}{{\sc r}e{\sc x}\xspace}
\newcommand{\imagenet}{{{\sc i}mage{\sc n}et}\xspace}
\newcommand{\dc}{{\sc dc}\xspace}

\newcommand{\xai}{XAI\xspace}
\newcommand{\commentout}[1]{}

\usepackage{mdframed}

\newenvironment{resframe}[1]
{\mdfsetup{
    frametitle={\colorbox{blue!5}{\space#1\space}},
    innertopmargin=8pt,
    frametitleaboveskip=-\ht\strutbox,
    frametitlealignment=\center,
    backgroundcolor=black!10,
    roundcorner=10pt,
    nobreak=true,
    }
  \begin{mdframed}%
  }
  {\end{mdframed}}

\usepackage{tabularx}
\newcolumntype{B}{p{3.5cm}}
\newcolumntype{s}{>{\raggedleft\arraybackslash $}p{1.3cm}<{$}}
\newcolumntype{q}{>{\raggedleft}p{1.7cm}<}

\newtheorem{definition}{Definition}

\begin{document}

\title{\texorpdfstring{I Am Big, You Are Little;\\I Am Right, You Are Wrong}{I Am Big, You Are Little; I Am Right, You Are Wrong}}

\author{David A Kelly \\
King's College London, UK\\
{\tt\small david.a.kelly@kcl.ac.uk}
\and 
Akchunya Chanchal\\
King's College London, UK\\
{\tt\small akchunya.chanchal@kcl.ac.uk}
\and 
Nathan Blake\\
King's College London, UK\\
{\tt\small nathan.blake@kcl.ac.uk}
}

\maketitle

\begin{abstract}
Machine learning for image classification is an active and rapidly developing field. With the proliferation of classifiers of different sizes and different architectures, the problem of choosing the right model becomes more and more important. While we can assess a model's classification accuracy statistically, our understanding of the way these models work is unfortunately limited. In order to gain insight into the decision-making process of different vision models, we propose using \emph{minimal sufficient pixels sets} to gauge a model's `concentration': the pixels that capture the essence of an image through the lens of the model. By comparing position, overlap, and size of sets of pixels, we identify that different architectures have statistically different concentration, in both size and position. In particular, ConvNext and EVA models differ markedly from the others. We also identify that images which are misclassified are associated with larger pixels sets than correct classifications. 
\end{abstract}

\section{Introduction}\label{sec:intro}
\begin{figure*}[t]%
    \centering
    \begin{subfigure}[b]{0.19\textwidth}
        \centering
        \includegraphics[scale=0.18]{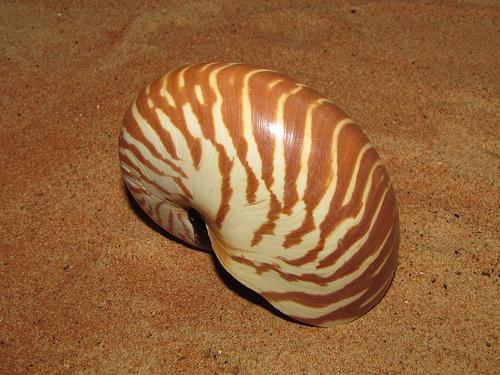}
        \caption{Original}
    \end{subfigure}
    \hfill
    \begin{subfigure}[b]{0.19\textwidth}
        \centering
        \includegraphics[scale=0.75]{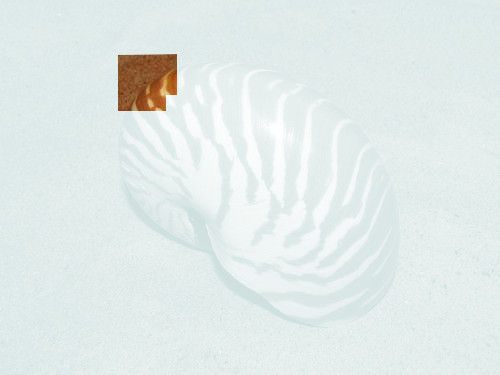}
        \caption{ConvNext-v2}
        \label{fig:seashellconv}
    \end{subfigure}
    \hfill
    \begin{subfigure}[b]{0.19\textwidth}
        \centering
        \includegraphics[scale=0.75]{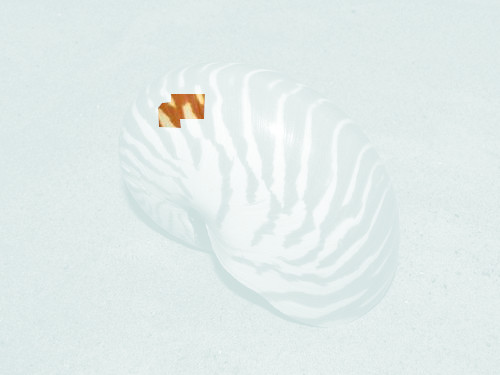}
        \caption{eva-v2-giant}
    \end{subfigure}
    \hfill
    \begin{subfigure}[b]{0.19\textwidth}
        \centering
        \includegraphics[scale=0.75]{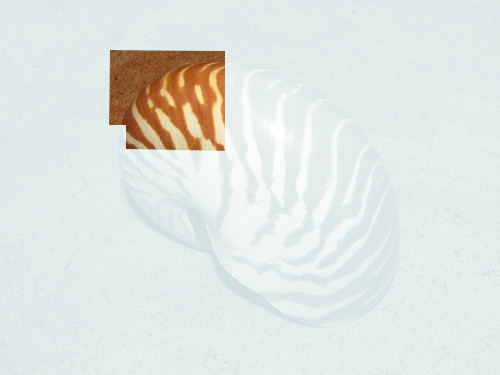}
        \caption{Inception-v4}\label{fig:seashellinception}
    \end{subfigure}
    \hfill
    \begin{subfigure}[b]{0.19\textwidth}
        \centering
        \includegraphics[scale=0.75]{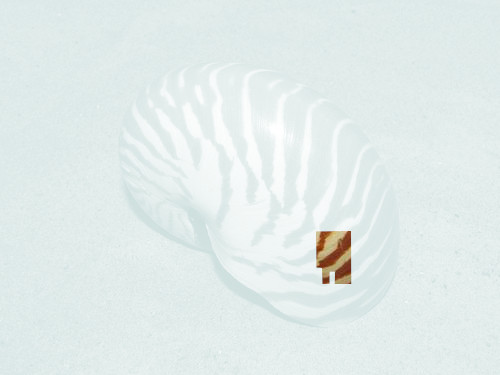}
        \caption{resnet152-a2}
        \label{fig:seashellresnet}
    \end{subfigure}
\caption{Minimal, sufficient pixels for an image classified as a seashell by different models. The varying size of the pixel sets is visually evident. ResNet152 uses a completely different region of the image, and very few pixels, for the classification.}%
\label{fig:seashell}
\end{figure*}

Neural networks are now a primary component of most AI systems, especially in computer vision. There are a plethora of studies comparing image classifiers, either within or across different architectures~\cite{NEURIPS2021_652cf383, NEURIPS2021_c404a5ad, mustapha2010comparison}. Such comparisons tend to focus on accuracy, precision, and robustness~\cite{Su_2018_ECCV}. These studies leave unanswered the question of what features or concepts models use to make their classifications. 

\citet{Jiang_2024_CVPR} investigate ``compositionality'', which they define as a conjunction of \emph{patches} that have high likelihood ratios for a particular classification. A patch is a fixed square in a grid constructed over the image. Combinations of patches are called \emph{minimally sufficient explanations} (MSEs), and are calculated using the explainable AI (\xai) tool,~\sag~\cite{sag}. As~\citet{chockler2023multiple} point out though, \sag uses a definition of explanation which is highly unusual. A combination of patches is considered sufficient if its classification likelihood is above a user-provided threshold. This threshold is a scalar of the model's confidence on an image and can be arbitrarily low. In effect, with a suitable threshold, \emph{any} combination of patches is sufficient for the desired classification. This sufficiency is unclear: it is certainly not always sufficient to guarantee that a patch combination provides enough information to (re)produce the original classification. Indeed, given an image with low model confidence, even a relatively high threshold is likely to see many accepted patch combinations which tell us nothing about the classification and very little about the model.

We start instead with the problem of `concentration': the smallest number of pixels required to get the original classification.
We use \rex\footnote{\url{https://github.com/ReX-XAI/ReX}}~\cite{CKKS24} to find \emph{minimal sufficient pixel sets}, or \mpss (see \Cref{fig:seashell}).
These have two important advantages over the MSEs of \sag. Firstly, they are not bound to a particular patch size. \sag breaks the image up into a grid of patches, so its MSE is minimal only in its combination of \emph{patches} and not in its combination of \emph{pixels}. \rex is not patch based, so an \mps is likely to contain fewer unnecessary pixels than an MSE\footnote{It is still only an approximation of true minimality however. See~\citet{CKKS24} for a full discussion.}. Secondly, a \rex \mps is sufficient to recreate the original (top) classification of the image. Instead of having a confidence threshold, \rex provides an approximately minimal, non patch-based, set of pixels which really are sufficient to get the desired classification from the model. 

By putting concentration onto a sound footing, other properties of models, such as compositionality --- a measure of diversity in model concentration --- can be studied more rigorously. We investigate concentration in terms of size, position and overlap of \mpss on $15$ different image classification models. The models cover $5$ different architectures and range in size from small Inception models to state-of-the-art transformers models with over a billion parameters. All models were fine-tuned on ImageNet. The goal of this comparison is to answer the following research questions:

\begin{description} 
    \item[RQ1] Do different models have \mpss of different sizes on the same image? 
     \item[RQ2] Do different models produce an \mps in different locations on the same input image?
     That is, are different models looking at different areas of the image for the classification?
     \item[RQ3] Do wrong classifications have larger or smaller \mpss than correct classifications? 
\end{description}

We find that larger models tend to produce smaller, more spatially distinct, minimal pixel sets. In particular, EVA and ConvNext are significantly different from the other studied models. This may indicate overfitting in these large models.

Although \rex is an \xai tool, in general we avoid using the word ``explanation'' for the rest of the paper. Our goal is not to demonstrate the utility of causal reasoning to human-centered explainability, 
but rather to capture something less subjective about how models process inputs and what features and concepts they rely upon. Indeed, given the surprisingly small size of many \mpss (see~\Cref{sec:eval}) it is not always obvious
how well they would act as explanations to a human. 

We also do not claim that a given \mps is the \emph{only} information relevant to the model when it makes its classification. Indeed, both~\cite{sag} and~\cite{chockler2023multiple} show that multiple explanations may exist in an image. The \mpss we consider are always the global maximum \mps (see \Cref{sec:method}): they are the pixels most causally responsible and show where the model was giving the majority of its concentration.

Due to the lack of space, full tables of results, models, benchmarks, and the experimental setup are submitted as a part of the supplementary material.

\section{Related Work}\label{sec:relwork}
Image classifiers have been developed since the 1980s, though it is only within the last two decades that computing power and data quantity have been sufficient to see them widely adopted~\cite{8016501}. Many studies compare models and architectures for accuracy and precision~\cite{WANG202161, 9687944, SHARMA2018377}. More unusually, Su~\etal~\cite{Su_2018_ECCV} compare $18$ different \imagenet models for robustness in the face of adversarial attacks. We investigate, however, the minimal pixels required for a model to reproduce its original classification. We conduct this investigation using a causal \xai tool, \rex~\cite{CKKS24}. We give an overview of \rex in~\Cref{sec:method}.

We use \rex because the causal definition of an explanation (\Cref{defn:simple-exp}) is suitable to our needs. It has the advantage over other \xai tools in that the ``explanation'' is tested modulo the model: a causal explanation is a minimal sufficient subset of pixels in the image such that they have the same classification as the overall image. As the model acts as its own oracle, a causal explanation is less dependent on human interpretation than, for example, Shapley values. \rex is a purely black-box tool.
Broadly, \xai tools split into white and black-box methods~\cite{whiteboxreview}. The white-box method \gradcam~\cite{CAM}, for example, has spawned extensions~\cite{camplusplus} and forms just a small part of the wide range of layer attribution methods available\footnote{See the captum library~\cite{captum} for a large selection of algorithms.}. All white-box methods require access to the internals of the model and need to be tailored to the specifics of the model architecture.

Black-box methods form a smaller family, but still exhibit a diversity of approaches. Among the more popular tools 
for image classifier explanations are \lime~\cite{lime}, \shap~\cite{shap}, and \rise~\cite{rise}. \lime builds a locally interpretable model by using perturbations of the image. 
\shap uses game-theoretic Shapley values to provide a heatmap of pixel contributions to the classification. \rise utilizes random occlusions of the image to discover pixel contributions to the classification. 
All of these tools provide some form of pixel ranking, but do not directly isolate those pixels sufficiently for a given classification.

\citet{Jiang_2024_CVPR} use the \xai tool \sag~\cite{sag} to compare the decision making mechanisms of transformers and CNNs. \sag generates multiple explanations for a given image. These explanations are given as ``patches'', similar in intent to our \mpss. However, the patch itself is of a fixed size and explanations are always combinations of these fixed size patches. \rex has no such limitation. Moreover, what \sag accepts as a minimal sufficient explanation is much more generous than what \rex accepts. \rex accepts the \mps if, and only if, the top class of the \mps is the same as the top class of the overall image: \sag instead uses a confidence threshold. This means that a \sag MSE might not be the top $1$ classification, but in fact might be in almost any position in the output tensor~\cite{chockler2023multiple}.

\section{Minimal Pixels Sets}\label{sec:method}

\begin{figure*}[t]
    \centering
    \begin{subfigure}{0.2\textwidth}
        \centering
        \includegraphics[scale=0.2]{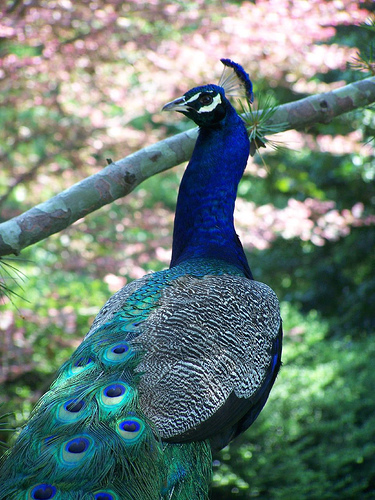}
        \caption{A peacock}\label{fig:peacock}
    \end{subfigure}
    \hfill
       \begin{subfigure}{0.2\textwidth}
        \centering
        \includegraphics[scale=0.23]{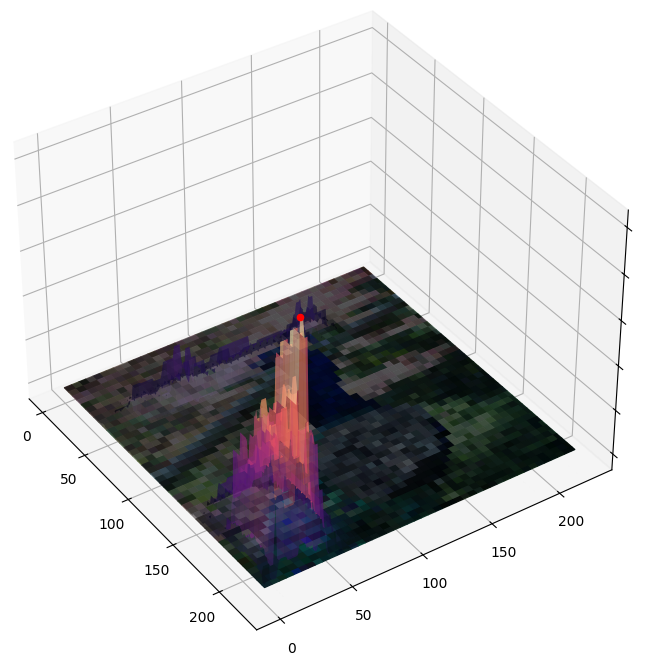}
        \caption{Responsibility map}\label{fig:peacock_map}
    \end{subfigure}
    \hfill
    \begin{subfigure}{0.2\textwidth}
        \centering
        \includegraphics[scale=0.45]{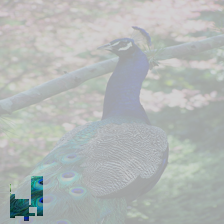}
        \caption{A peacock \mps}\label{fig:peacock_mps}
    \end{subfigure}
      \hfill
    \begin{subfigure}{0.2\textwidth}
        \centering
        \includegraphics[scale=0.45]{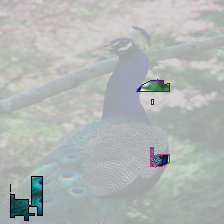}
        \caption{Other peacock \mpss}\label{fig:peacock_multi}
    \end{subfigure}
    
    \caption{The various stages of calculating an \mps. We start with an image,~\Cref{fig:peacock}. \rex produces a responsibility landscape, (\Cref{fig:peacock_map}), which shows an approximation of the causal responsibility of each pixel towards the classification ``peacock''. This map is used to rank pixels, which are then introduced over the baseline value until the pixels reproduce the classification. The result is an \mps (\Cref{fig:peacock_mps}). \rex can also find multiple different \mpss (\Cref{fig:peacock_multi}). In this paper, we investigate the best \mps, the one ranked most highly in~\Cref{fig:peacock_map}.}\label{fig:rex}
\end{figure*}

We use \rex~\cite{CKKS24} to construct
sufficient pixel sets in image classification. The reader is referred 
to~\citet{Hal19} for a detailed formal overview and more information on actual causality and responsibility. Broadly speaking, \rex views a model as a black-box causal model in the Halpern-Pearl~\cite{HP05a} sense of the word,
with its inputs being the individual pixels of an image. The variables are
defined as Boolean, with the values being the original color and a baseline
value. The relevant general definitions are given in~\cite{Hal19}. 
Here we only present the definition of explanation for image classification.

\begin{definition}[CKS explanation for image classification]\label{defn:simple-exp}
An explanation in image classification is a minimal subset of pixels of a
given input image that is sufficient for the model to classify the image,
where ``sufficient'' is defined as containing only this subset of pixels
from the original image, with the other pixels set to a baseline value.
\end{definition}
\noindent

A recent paper by~\citet{CH24} proves that under the same simplifying assumptions that \rex uses, \Cref{defn:simple-exp} is equivalent to the definition of explanation in actual causality~\cite{Hal19}. 
\citet{CH24} observe that the precise computation of an explanation in our setting is intractable, as the problem is equivalent to 
an earlier definition of explanations in binary causal models, which is DP-complete~\cite{EL04} (DP is the class of 
languages that are an intersection of a language in NP and a language in co-NP and contains, in
particular, the languages of unique solutions to NP-complete problems~\cite{Pap84}).

\paragraph*{Algorithmic Overview} We present a simplified description of the \rex algorithm. See \citet{CKKS24} for a complete description. 
\rex starts by dividing an image into $4$ parts. We call each part a superpixel. \rex creates \emph{mutants} of the original image by covering all combinations of superpixels with a baseline value. By default, this baseline value is $0$.
These combinations are tested against the model and sorted between those which satisfy the required classification and those which do not. 
Note that some superpixels may appear in both passing and failing mutants, depending on the exact combination. 
Causal responsibility is distributed over the different superpixels,
where responsibility is a quantitative measure of causality and, broadly speaking, measures the amount of causal influence on the classification~\citep{CH04}. Passing superpixel combinations are further refined into smaller superpixel combinations and tested using the same procedure.

In this way, mutant generation is iteratively guided by the model, and responsibility tends to concentrate on those pixels (not superpixels) which occur most frequently in ``good'' combinations. This procedure is repeated many times, starting from a different random partition of the image. After a number of iterations of the algorithm, by default $20$, \rex produces a responsibility landscape not dissimilar to~\Cref{fig:peacock_map}. One can immediately see that causal responsibility is peaked over one small part of the image. This is not true in general, especially if multiple explanations are present (\cite{chockler2023multiple}).

This landscape is not itself an explanation, or \mps, according to~\Cref{defn:simple-exp}. \rex still needs to identify, from the landscape, those pixels which are minimal and sufficient for the classification ``peacock''.
The pixels are ranked in the order of their \emph{responsibility} for the classification, We note, however, that the construction of the ranked list is intractable as well (NP-complete), 
even  in the special case of image classification, rather than the general definition of responsibility by~\cite{CH04}. Hence, \rex's ranking is based on the approximate degree of responsibility.
As exact \mps computation is intractable, \rex's output is (approximately) minimal, but not necessarily the minimum. As \rex adds pixels according to the responsibility ranking, the first discovered \mps is the one with highest responsibility \mps (\Cref{fig:peacock_mps}).
It is guaranteed to be sufficient to obtain the original classification, against the provided baseline. These sufficient sets may sometimes have very low model confidence. We do not address here the issue of whether there is a threshold of 
model confidence below which we can no longer give credence to the results. 
Pixel sets based on \Cref{defn:simple-exp} have the clear advantage of being amenable to a comparison
using standard approaches, such as the S\o{}rensen–Dice coefficient (\dc), which is used to gauge the similarity of any two samples~\cite{dic45,sor48}, and the Hausdorff distance~\cite{hausdorff}.

\paragraph*{Out of distribution (ood) mutants} We use \rex with its default baseline masking value of $0$. This masking value is applied after processing the image, and therefore does not necessarily correspond to black. It does mean that virtually every mutant created by \rex results is an \emph{ood} image. The fact that the models are still able to classify the image in a ``sane'' fashion is a good indicator of their ability to cope with this type of \emph{ood} image. As \cite{CKKS24,blake2023mrxai} show, \rex \mpss are almost always inside, or overlapping, a human-provided segmentation, indicating that the \mps is in a reasonable position in the image. 
As we shall see, the size of a pixel set seems to be a good indicator of how robust a model is to \emph{ood} images, at least of the sort produced by \rex. 

\subsection{Models and architectures}\label{subsec:architecture}
We consider the following architectures in our study, with each architecture represented by three models. 

Inception~\cite{inception-v3} models are a family of networks based on CNN classifiers. 
Rather than simply stacking convolutional layers to achieve better performance, Inception models are more complex and utilize a large body of different optimizations and heuristics. Fundamentally, they work by having multiple filters of different sizes at each level of the network. This makes the network wider rather than deeper.

ResNet~\cite{ResNet} (\emph{Residual Network}) architecture was introduced to solve the vanishing/exploding gradient problem common to large CNNs. ResNet uses \emph{residual blocks}, where we skip certain intermediate layers when connecting two nodes. A ResNet is a stack of residual blocks. ConvNext~\cite{Liu_2022_CVPR} is a modernization of the standard ResNet architecture to bring it closer to the design of vision transformer models. The authors report that ConvNext models are competitive with vision transformers on both accuracy and scalability. 

ViT~\cite{Dosovitskiy2020AnII} (\emph{Vision Transformer}) is an image classification model which uses the encoder-only transformer architecture over patches of an image. 
An image is turned into a vector suitable for a standard transformer model by dividing it up into fixed-sized patches, which are then linearly embedded along with partition information. 
EVA~\cite{Fang_2023_CVPR} is a ViT pre-trained to allow the model to scale to a very large number of parameters, potentially over $1$ billion.

\section{Evaluation}\label{sec:eval}
\begin{table*}[t]
    \centering
    \begin{tabularx}{0.85\linewidth}{B|s|s|s|s||s|s}
    \multicolumn{1}{c}{} & \multicolumn{4}{|c}{Average} & \multicolumn{2}{|c}{} \\
    \cmidrule(lr){2-5} 
    \multicolumn{1}{c}{Model} & \multicolumn{1}{|c}{Area} & \multicolumn{1}{|c}{Correct} & \multicolumn{1}{|c}{Incorrect} & \multicolumn{1}{|c}{Test} & \multicolumn{1}{|c|}{Mean} & \multicolumn{1}{c}{Accuracy} \\
    \midrule
    \midrule
    ConvNext-V2 Large  &  0.081  & 0.075 & 0.122 & 0.078 & 0.089 & 0.880\\
    ConvNext-V2 Huge v2  &  0.065  & 0.063 & 0.082 & 0.061 & 0.068 & 0.894\\
    ConvNext-V2 Huge v1  &  0.05  & 0.048 & 0.061 & 0.05 & 0.052 & 0.890\\
    EVA-02 Large V1  &  0.066  & 0.064 & 0.082 & 0.068 & 0.07 & 0.890\\
    EVA-02 Large V2  &  0.07  & 0.068 & 0.09 & 0.065 & 0.073 & 0.894\\
    EVA Giant &  0.054  & 0.052 & 0.065 & 0.052 & 0.0558 & 0.894\\
    Inception-ResNet V2  &  0.25  & 0.246 & 0.265 & 0.253 & 0.254 & 0.814\\
    Inception V3  &  0.239 & 0.231 & 0.271 & 0.245 & 0.247 & 0.800\\
    Inception V4  &  0.23  & 0.224 & 0.261 & 0.243 & 0.239 & 0.840\\
    ResNet 152-B A1  &  0.142  & 0.137 & 0.166 & 0.139 & 0.146 & 0.828\\
    ResNet 152-B A2  &  0.144  & 0.136 & 0.187 & 0.149 & 0.154 & 0.842\\
    ResNet152-D &  0.134  & 0.13 & 0.155 & 0.127 & 0.137 & 0.828\\
    ViT Large  &  0.099  & 0.098 & 0.111 & 0.089 & 0.099 & 0.900 \\
    ViT Huge V1 &  0.156  & 0.154 & 0.17 & 0.152 & 0.158 & 0.882\\
    ViT Huge V2 &  0.103  & 0.102 & 0.113 & 0.095 & 0.103 & 0.872\\
    \bottomrule
    \end{tabularx}
    \caption{Average size of \mps as percentage of entire image. \emph{Correct} is average area for correct classifications and \emph{Incorrect} for incorrect classifications. We do not have ground truth labels for \emph{Test} so present average area without differentiation.
    The difference between the Inception models and ConvNext in particular is remarkable, being $3.6\times$ larger on average.~\emph{Accuracy} is the accuracy of the model upon the $500$ images for which we have ground truth labels.}
    \label{tab:average_area}
\end{table*}

We compare $15$ different models, drawn from $5$ different architectural classes. All experiments were run on Ubuntu 20.04 LTS on a single A-100 GPU. \rex was run with the same hyperparameters and same random seed for all experiments. 
For the models, we used the latest available pretrained IN-1k weights, dated February 22nd 2024, via the PyTorch Image Models (Timm) package. All the models were then converted to the ONNX format, using ONNX v1.15 and opset 20. The model was then run under ONNX runtime 1.17 for inference when performing all the experiments. This was done due to the prevelance of deploying models using ONNX in production environments and hence, allowed for replicating real-world conditions as closely as possible.

\subsection{Experimental Design}\label{subsec:design}
The complete list of all the different models is presented in~\Cref{tab:average_area}. Two of these architectures are transformer-based (ViT and EVA), and the rest are convolutional. Each model has a different input size and a different number of internal parameters. They also differ in their pre-training data set. All models are fine-tuned on \imagenet-1k. To further validate our results, we have also fine-tuned the models on the Caltech-256 dataset to compare
against the results obtained on \imagenet-1k. Due to the cost of running the experiments, we selected $500$ images uniformly at random from the validation set of \imagenet-1k. We similarly selected $500$ from the test set to produce a total of $1000$ images. 

\subsection{Statistical analysis}\label{subsec:stat}
The subsequent analyses are based on the ratio of the \mps size to the size of the whole image (as different models accept different, fixed, input sizes). We calculate this ratio on different partitions of \imagenet-1k, the validation and test set, the former of which has ground truth labels. 
Our first null hypothesis is that there is no statistical difference in \mps size between architectures on the test set data. Our second null hypothesis is that incorrect classifications have the same size \mps as correct classifications (see~\Cref{tab:average_area}), tested on the validation set only.
By having two random samples we mitigate against type $1$ errors, rejecting of the null hypothesis when it is in fact true. Additionally, when we apply these post hoc tests on our data, we use the \emph{Bonferroni correction}~\cite{Bon36} to counteract the problem of applying multiple hypothesis tests to the same data.

We use the Kruskal-Wallis $H$ test~\cite{kruscal} and Friedman test~\cite{siegel1988nonparametric}, both non-parametric tests for non-normal data. The former is used for cross-architecture analysis. The latter is used to detect differences between different treatments of the same data. Because our data is matched (\ie \mpss were extracted from the same images), we can use this test to detect intra-architecture differences in \mps size. We treat the threshold of statistical significance as $p < 0.01$.

We additionally calculate two sets of \dc and Hausdorff values, for comparing the outputs of models across architectures and of the same architecture. For the cross-architecture analysis, we select the best-performing model from each architecture class based on accuracy scores on the randomly selected samples from \imagenet-1k and Caltech-256 (\Cref{tab:acc}). We then find the intersections of both correct and incorrect classifications (separately) for our best performing models. We calculate our measures on these subsets. As we do not have labels for the test set of \imagenet-1k, we use the majority vote between models as a proxy for correctness. We use a similar approach for intra-architecture measures, comparing all models of a given architecture.

\subsection{Results}\label{subsec:results}
In this section we present the analysis of our experiments. All of our analyses are over sets of \mpss, obtained from different models on the same images.
\Cref{fig:seashell} is an illustration of such a set, for a seashell classification. We can see that the \mps in~\Cref{fig:seashellresnet} is in a completely different location
from other explanations, having a \dc of $0$ and a Hausdorff distance of approximately $156$ from~\Cref{fig:seashellconv}. All other \mpss in this example cluster 
around the upper part of the shell. \rex was used with the same random seed for all runs, meaning that all \emph{initial} superpixels were the same. Refinement of superpixels is guided by the model under test. The fact that ResNet152 has an \mps in a significantly different area indicates that causal responsibility was distributed completely differently from the other models.

We focus on the results for \imagenet in this section. We verified the \imagenet-1K results using a set of $50$ randomly sampled images from CalTech-256 as a sanity check on our main results. This returns similar findings, with a statistically significant difference between architectures ($p < 0.01$), though evidence of inter-architecture differences are not found for the EVA, Inception or ResNet models at $p < 0.01$. The sample size of CalTech-256 was much smaller however.

\begin{resframe}{RQ1: different \mps size}
There are statistically significant differences in \mps size both across and inside different architectures. The Kruskal-Wallis $H$ test indicates the probability of the null hypothesis being correct at $p < 0.01$.
\end{resframe}

\begin{figure}[t] 
    \centering
    \includegraphics[scale=0.28]{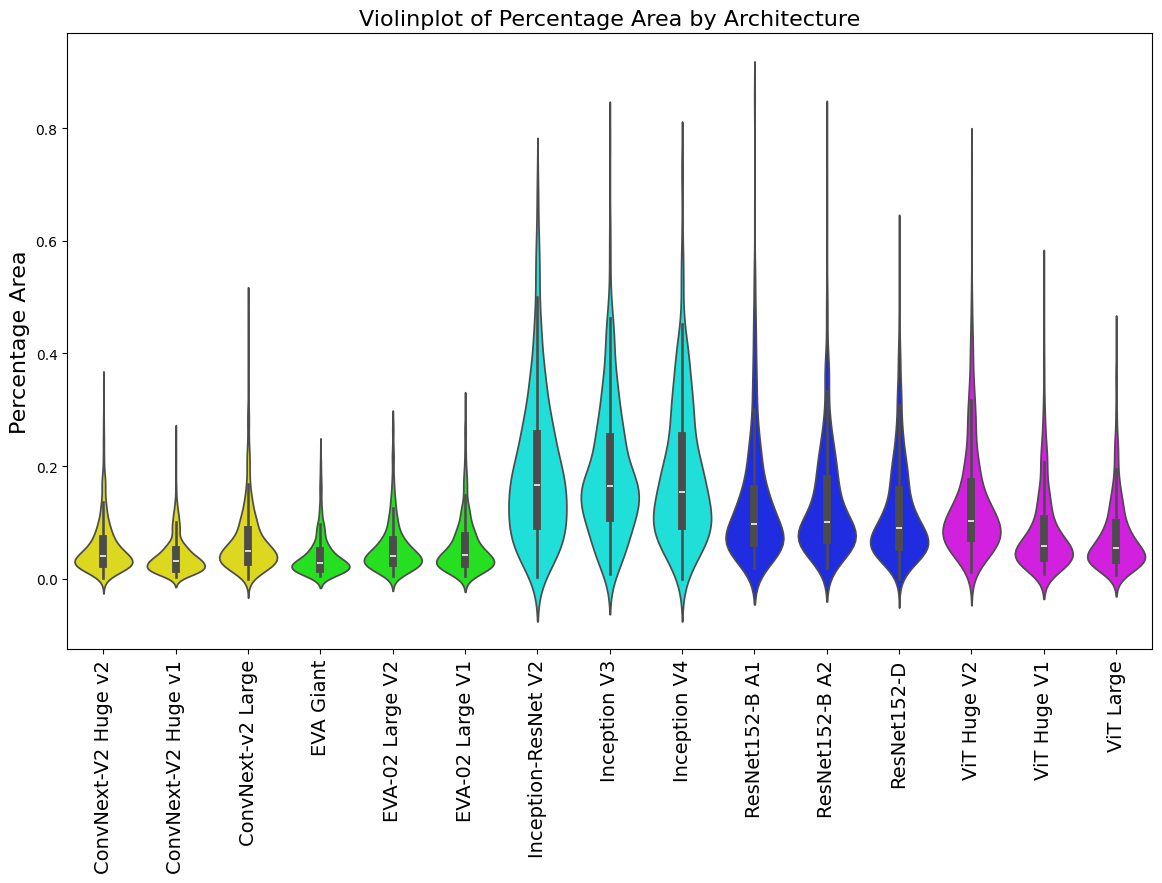}
    \caption{Violin plot of \mps size as a ratio to overall image size. Color coordinated by architecture with yellow: ConvNnet, green: EVA-02, cyan: Inception, dark blue: ResNet and purple: ViT. A Kruskal–Wallis $H$ test indicates a significant difference amongst the architectures: $H(4) = 1176.134, p < 0.001$. To determine whether there is a difference between the models within each architecture we apply the Friedmans test results and find that there is a difference, except for the Inception models (p=0.36).}
\label{violin_by_architecture}
\end{figure}

Both the \dc and Hausdorff scores for our top models demonstrate that \mpss of models with different architectures are not based on the same set
of pixels. There is considerable variety of \mps size and location depending on the model.
\Cref{tab:average_area} presents the ratios of the size of the \mps with respect to the size of the input image, for all pixel sets and separately for the subsets of correct and of incorrect classifications. The Kruskal-Wallis $H$ test indicates that, based on the \imagenet-1k dataset, the size of pixel sets found by \rex does differ by model architecture ($p < 0.01$). The Inception models, in particular, return larger pixel sets (see \Cref{fig:seashellinception} for a typical example).
The Friedman test further demonstrated a significant difference in \mps size even within architectures, except for the Inception models (ConvNext, $p < 0.01$; EVA $p < 0.01$; Inception, $p > 0.01$; ResNet $p < 0.01$; ViT, $p < 0.01$). \Cref{violin_by_architecture} shows a violin plot of all models and their \mps size, with architecture coordinated by color. It is immediately clear the the ConvNext and EVA models are quite different from the other architectures.

\begin{resframe}{RQ2: \mps location}
Both the \dc and Hausdorff scores for our top models demonstrate that \mpss do not group around the same set of pixels. There is a considerable variety of \mps location. 
\end{resframe}

\begin{table}[t]
    \centering
    \begin{tabular}{l||rrrrr}
        \multicolumn{1}{c||}{Model} & \multicolumn{1}{C}{\small{EVA Giant}} & 
        \multicolumn{1}{C}{\small{Conv Next}} & \multicolumn{1}{C}{\small{ViT Large}} & 
        \multicolumn{1}{C}{\small{ResNet 152}} & \multicolumn{1}{C}{\small{Inception}} \\
        \midrule \midrule
        \small{EVA Giant} & 1.0 & 0.287 & 0.253 & 0.165 & 0.141 \\
        \small{ConvNext} & 0.287 & 1.0 & 0.304 & 0.162 & 0.163 \\
        \small{ViT Large} & 0.253 & 0.304 & 1.0 & 0.232 & 0.225 \\
        \small{ResNet152} & 0.165 & 0.162 & 0.232 & 1.0 & 0.282 \\
        \small{Inception} & 0.141 & 0.163 & 0.225 & 0.282 & 1.0 \\
        \bottomrule
    \end{tabular}
    \caption{Average \dc values for pixel sets of best performing models across architectures on \imagenet-1k validation subset of correctly classified images.} 
    \label{tab:dice}
\end{table}

\Cref{tab:dice} shows the average \dc for the top performing models. The average overlap is in general quite low, indicating that \mpss across models do not share a large number of pixels.
\Cref{tab:hausdorff} shows the average Hausdorff distance across our best performing models. Inception V4 consistently finds pixel sets in different locations from these for the other models. The closest to these of Inception V4 are produced by ResNet152.

\begin{figure*}[t]%
    \centering
    \begin{subfigure}[b]{0.3\textwidth}
        \centering
        \includegraphics[scale=0.18]{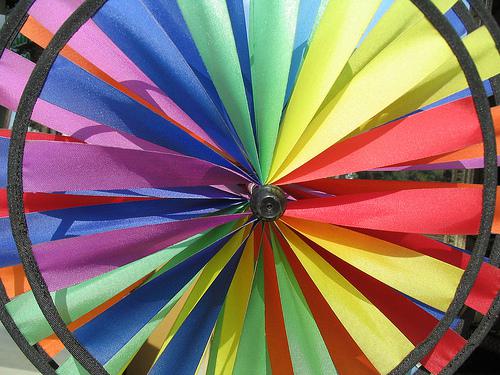}
        \caption{Original}
    \end{subfigure}
    \begin{subfigure}[b]{0.3\textwidth}
        \centering
        \includegraphics[scale=0.75]{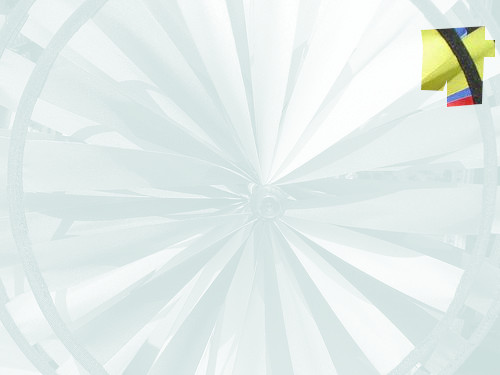}
        \caption{CN-v2}
        \label{fig:pinconv}
    \end{subfigure}
    \begin{subfigure}[b]{0.3\textwidth}
        \centering
        \includegraphics[scale=0.75]{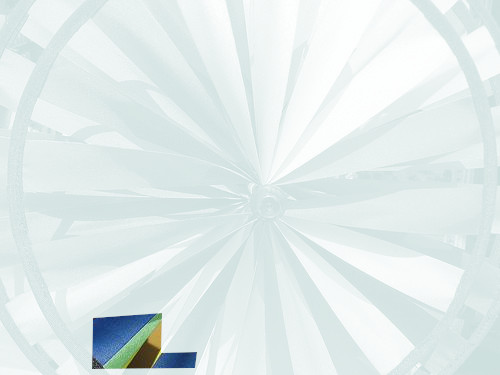}
        \caption{EVA}
    \end{subfigure}
    \begin{subfigure}[b]{0.3\textwidth}
        \centering
        \includegraphics[scale=0.75]{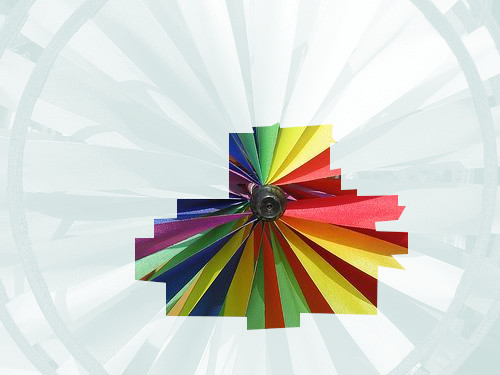}
        \caption{Inception}%
        \label{fig:pininception}
    \end{subfigure}
    \begin{subfigure}[b]{0.3\textwidth}
        \centering
        \includegraphics[scale=0.75]{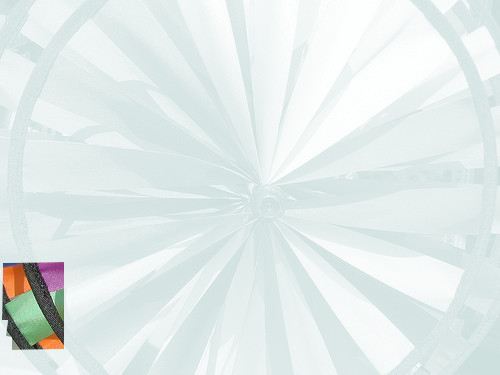}
        \caption{ResNet152}
        \label{fig:pinresnet}
    \end{subfigure}
    \begin{subfigure}[b]{0.3\textwidth}
        \centering
        \includegraphics[scale=0.75]{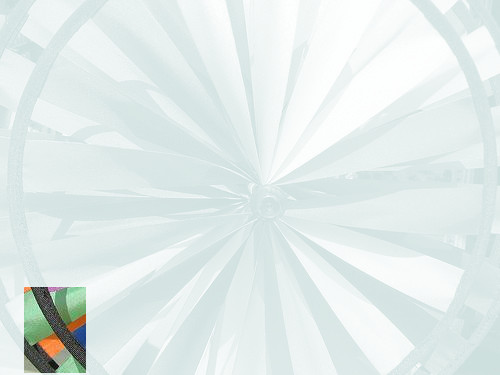}
        \caption{ViT-Large}
        \label{fig:pinvit}
    \end{subfigure}
\caption{A pinwheel. This image has the highest Hausdorff distance for any set of \mpss over our best performing models. While a high Hausdorff distance does not necessarily imply \emph{no} pixel overlap, in this case only \Cref{fig:pinresnet} and \Cref{fig:pinvit} have any pixels in common.}%
\label{fig:pinwheel}
\end{figure*}

\begin{table}[t]
    \centering
   \begin{tabular}{l||rrrrr}
        \multicolumn{1}{c||}{Model} & \multicolumn{1}{C}{\small{EVA Giant}} & 
        \multicolumn{1}{C}{\small{Conv Next}} & \multicolumn{1}{C}{\small{ViT Large}} & 
        \multicolumn{1}{C}{\small{ResNet 152}} & \multicolumn{1}{C}{\small{Inception}} \\
        \midrule \midrule
        \small{EVA Giant} & 0.0 & 99.5 & 98.6 & 85.5 & 139.2 \\
        \small{ConvNext} & 99.5 & 0.0 & 95.6 & 89.9 & 139.4 \\
        \small{ViT Large} & 98.6 & 95.6 & 0.0 & 78.9 & 121.5 \\
        \small{ResNet152} & 85.5 & 89.9 & 78.96 & 0.0 & 88.4 \\
        \small{Inception} & 139.2 & 139.4 & 121.4 & 88.4 & 0.0 \\
        \bottomrule
    \end{tabular}
    \caption{Average Hausdorff coefficient values for explanations of best performing models across architectures on \imagenet-1k validation subset collectively classified correctly.}
    \label{tab:hausdorff}
\end{table}

\begin{resframe}{RQ3: incorrect \mps size}
There is a statistically significant difference between \mps sizes for correct and for incorrect classifications. The effect size is small, however, with only an average increase in area of $2.6$\%.
\end{resframe}

\begin{table}[t]
\begin{tabular}{l|r|r}
Model               & \multicolumn{1}{l|}{Imagenet-1K} & \multicolumn{1}{l}{Caltech-256} \\ 
\midrule
\midrule
EVA-02 Large V1     & 0.890    & 1.00    \\ \hline
EVA-02 Large V2     & 0.890    & 0.98  \\ \hline
EVA Giant           & 0.894    & 0.94     \\ \hline
ConvNext-V2 Huge v1 & 0.890   & 0.98   \\ \hline
ConvNext-V2 Huge v2 & 0.894   & 0.98  \\ \hline
ConvNext-V2 Large   & 0.880    & 1.00  \\ \hline
ViT Huge V1         & 0.882  & 0.98  \\ \hline
ViT Huge V2         & 0.872   & 0.98   \\ \hline
ViT Large           & 0.900  & 1.00  \\ \hline
ResNet152-B A1      & 0.828   & 0.92  \\ \hline
ResNet152-B A2      & 0.842   & 0.94  \\ \hline
ResNet152-D         & 0.828  & 0.94   \\ \hline
Inception V3        & 0.800  & 0.92  \\ \hline
Inception V4        & 0.840   & 0.82  \\ \hline
Inception-ResNet V2 & 0.814   & 0.88  \\ \hline
\bottomrule
\end{tabular}
\caption{Accuracy of different models on the randomly selected \imagenet-1k (validation set) and Caltech-256 samples.}
\label{tab:acc}
\end{table}

To assess whether the correctness of a model's output with respect to the ground truth label, or \emph{fidelity}, has an effect on the \mps size, 
we partitioned the classifications of the \imagenet validation set into correct and incorrect subsets. As evident from table~\Cref{tab:acc}, 
different models differ in their accuracy by as much as $10$\%, indicating a notable confounding effect. To control for this, we applied a mixed linear model, which demonstrated that having the incorrect classification increased the explanation size by 2.6\% (standard error $0.4\%, p < 0.01$). A similar effect is seen in the CalTech-256 data, with an estimated increase in size of $3.3$\% (standard error $0.01\%, p < 0.01$). 
While statistically significant, practical implications may be limited; that is something that needs to be determined qualitatively, and may change by domain area.

\begin{figure*}[t]%
    \centering
    \begin{subfigure}[b]{0.3\textwidth}
        \centering
        \includegraphics[scale=0.24]{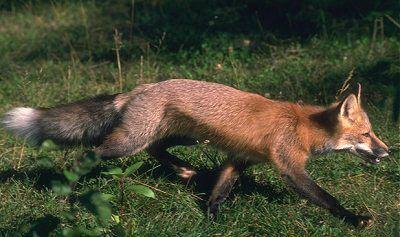}
        \caption{Original}
    \end{subfigure}
    \begin{subfigure}[b]{0.3\textwidth}
        \centering
        \includegraphics{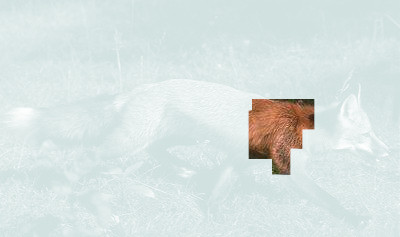}
        \caption{CN-v2}
        \label{fig:foxconv}
    \end{subfigure}
    \begin{subfigure}[b]{0.3\textwidth}
        \centering
        \includegraphics{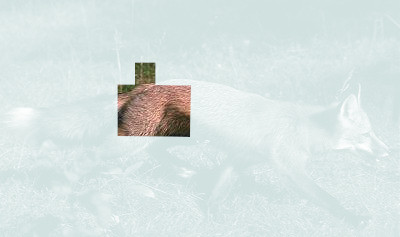}
        \caption{EVA}
    \end{subfigure}
    \begin{subfigure}[b]{0.3\textwidth}
        \centering
        \includegraphics{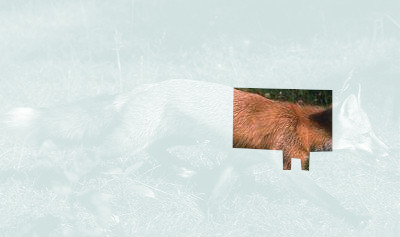}
        \caption{Inception}
    \end{subfigure}
    \begin{subfigure}[b]{0.3\textwidth}
        \centering
        \includegraphics{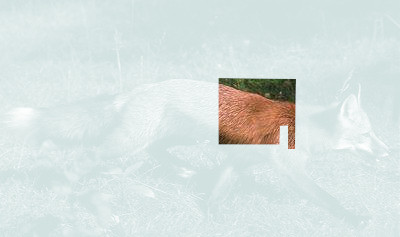}
        \caption{ResNet152}
        \label{fig:foxresnet}
    \end{subfigure}
    \begin{subfigure}[b]{0.3\textwidth}
        \centering
        \includegraphics{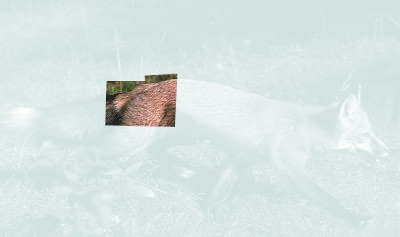}
        \caption{ViT-Large}
        \label{fig:foxvit}
    \end{subfigure}
\caption{The label for this image is ``grey fox''. All of the best performing models classify it as ``hyena''. One of the salient features for a human classifier, the tail, is completely missing from the \mpss.}%
\label{fig:fox}
\end{figure*}

\begin{figure*}[t]%
    \centering
    \begin{subfigure}[b]{0.24\textwidth}
        \centering
        \includegraphics[scale=0.19]{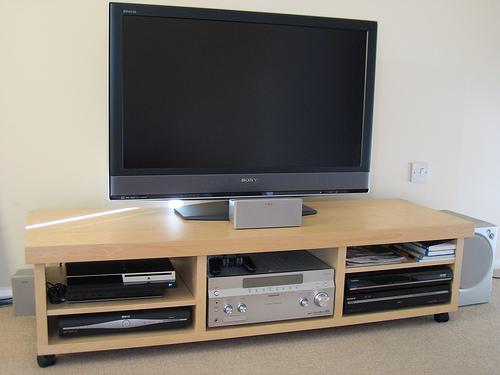}
        \caption{Original}
    \end{subfigure}
    \hfill
    \begin{subfigure}[b]{0.24\textwidth}
        \centering
        \includegraphics[scale=0.8]{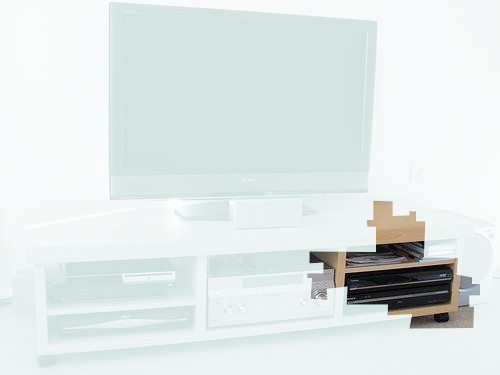}
        \caption{CN-Large}\label{fig:tvcnl}
    \end{subfigure}
    \hfill
    \begin{subfigure}[b]{0.24\textwidth}
        \centering
        \includegraphics[scale=0.8]{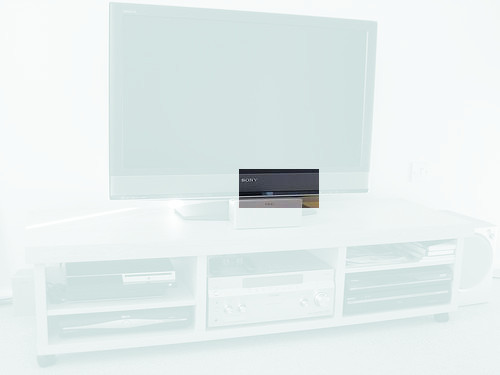}
        \caption{CN-Huge}\label{fig:tvcnh}
    \end{subfigure}
    \hfill
    \begin{subfigure}[b]{0.24\textwidth}
        \centering
        \includegraphics[scale=0.8]{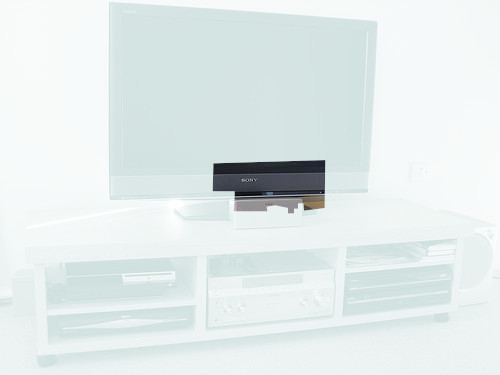}
        \caption{EVA}\label{fig:tveva}
    \end{subfigure}
\caption{The ground truth label for this image is $598$ ``home theater''. Only the EVA model (\Cref{fig:tveva}) produced the same classification. The two ConvNext models, \Cref{fig:tvcnl} and \Cref{fig:tvcnh} both classify the image as $851$, ``television''. It is interesting to note that the \mpss for ConvNext-Huge and EVA are very similar, though the classifications are different.}%
\label{fig:tv}
\end{figure*}

\Cref{fig:fox} shows an example of model \emph{infidelity} to the ground truth label. Its label in \imagenet is ``grey fox''. The best performing models all label the image as ``hyena''. The \mpss cluster around the center and rear of the body. Of note is that one of the salient distinguishing features of the grey fox to humans, the black stripe down its tail which is also tipped with black, is completely absent from the \mpss. 
Conversely, \Cref{fig:tv} is an example of mixed fidelity, where some models are faithful to the ground truth label and some are not. The disagreement in this case is logical, as there is potential ambiguity between the ``home theater'' and ``television''. It is interesting to note the inconsistency of the disagreement, with both~\Cref{fig:tvcnh} and~\Cref{fig:tveva} having \mpss of similar size and location, but different classifications.

\subsection{Discussion}\label{subsec:discussion}
\rex mutants are produced using a baseline masking value, as opposed to the blur that \sag uses. For an \mps to be very small, the model must accept a relatively large number of images which are fundamentally out of distribution. As \rex uses a process of iterative refinement for mutant creation and usually quits only when the mutants are no longer classified appropriately, it would seem that some architectures are quite happy to make classifications based on very little ``information''. These very small \mpss (see~\Cref{fig:pinconv}) may indicate overfitting. Conversely, models which produce large \mpss, such as Inception, are less able to understand these \emph{ood} images. It may be the case that different models are more sensitive to different types of \emph{ood} images (\ie the use of different baselines, or different types of blurring). 

Our results highlight the importance of taking \mps characteristics into account when selecting an appropriate model for a given task, rather than relying purely on other performance metrics. The results demonstrate that large models, which are pre-trained on larger corpora of data, on average, tend to utilize very little of the available input (5.4\% in the case of the 1 billion parameter EVA-Giant model). They are also highly confident in their predictions, highlighting the ``myopic'' tendencies of such models. This raises questions regarding the safety of these models, especially in high-stakes environments, such as healthcare, autonomous navigation or quality assurance.

\mpss provide a possible solution for mitigating some of these concerns. As there is a significant statistical difference between the size of the \mpss of a given model when it classifies an example correctly or incorrectly, this can be used as an additional check. Such a check would take place post-classification, to determine if the model's decision about the example is in the range of the \mpss of previously encountered correctly or incorrectly classified examples.

As \imagenet-1K labels are slightly ``noisy''~\cite{WDDBB22,RLI21,FLAWS25}, more work is needed to discover the impact of incorrect human labeling on the size of \mpss. While the problem of \imagenet-1K labels does not affect the size or positioning of \mpss, incorrect human labeling may have an effect on our analysis of \mps size for correct and incorrect classification. With that said, our results indicate that \mpss are too myopic in general, whether they are correctly classified or not.

\section{Conclusions and Future Work}\label{sec:conclusion}
In this paper, we present a large scale study on the comparison of \mpss on \imagenet-1k images. We demonstrated comprehensive experimental results on
$15$ different models across $5$ different model architectures. We used a state-of-the-art \xai tool, \rex, based on actual causality, to generate \mpss 
across the \imagenet-1k dataset. Using formal statistical methods, we demonstrated that there are statistically significant differences in \mps size and location
between different architectures and within the same architecture. 
We additionally showed that there is a small, but statistically significant, increase in explanation size for classifications which do not agree with the ground truth label.

While we are not focused on explainability, the question remains as to whether an \mps is a good explanation for a human. The answer to this question is ultimately qualitative: the datasets examined do not have human-annotated explanations, so do not allow a quantitative comparison between an \mps and a human explanation. Even if we had such information, models often base their classification on different features than humans do, 
as demonstrated in \Cref{fig:fox}. 

Our results find that different models have different degrees of confidence in their \mpss; some model types remain very confident in their classifications, even when the \mps itself is small. We will extend this investigation with the effect of model confidence on the size of \mpss. It is interesting to observe the presence of at least $2$ entirely disjoint explanations in~\Cref{fig:seashell}. This multiplicity of explanations is precisely the object of study of~\citet{Jiang_2024_CVPR}. Applying a similar methodology as used in this study to look at multiple explanations across multiple architectures may reveal new and interesting insights into model behavior.

\paragraph*{Acknowledgements}
The authors were supported in part by 
CHAI---the EPSRC Hub for Causality in Healthcare AI with Real Data (EP/Y028856/1).

\newpage
\bibliographystyle{ieeenat_fullname}
\bibliography{all}

\end{document}